# Multistep feature aggregation framework for salient object detection

Xiaogang Liu[a]   Shuang Song[b]

[a] School of Computer Science and Technology, Nanjing Tech University, Nanjing City, Jiangsu Province, 211800, PR China
e-mail: wtdf0201@163.com

[b] School of Computer Science and Technology, Nanjing Tech University, Nanjing City, Jiangsu Province, 211800, PR China
e-mail: songshuang@njtech.edu.cn

## Abstract

Recent works on salient object detection have made use of multi-scale features in a way such that high-level features and low-level features can collaborate in locating salient objects. Many of the previous methods have achieved great performance in salient object detection. By merging the high-level and low-level features, a large number of feature information can be extracted. Generally, they are doing these in a one-way framework, and interweaving the variable features all the way to the final feature output. Which may cause some blurring or inaccurate localization of saliency maps. To overcome these difficulties, we introduce a multistep feature aggregation (MSFA) framework for salient object detection, which is composed of three modules, including the Diverse Reception (DR) module, multiscale interaction (MSI) module and Feature Enhancement (FE) module to accomplish better multi-level feature fusion. Experimental results on six benchmark datasets demonstrate that MSFA achieves state-of-the-art performance.

**I. Introduction**

Salient object detection (SOD) is an important task in many computer vision fields, which tries to simulate the human visual system to distinguish the most visually obvious regions. It has been proved to play a useful role in many computer vision tasks, such as semantic segmentation, image retrieval, video abstraction, texture smoothing, and so on. In complicated scenarios, it still remains challenging to detect salient objects accurately [1–6].

Low-level cues and hand-crafted low-level features are usually used in traditional SOD methods. Color prior, central prior and boundary prior are some of the widely used low-level hand-crafted features [7–12]. However, due to the various scales of salient regions, these methods are limited by the absence of necessary high-level semantic information, and have difficulty in consistently detecting salient regions. In recent years, convolutional neural networks (CNNs) have been widely used in various visual tasks, and achieved great success because of their strong capability in feature extraction. CNN-based models employ multiscale learning to leverage low-level local features with rich details and high-level global representations for locating salient objects and discovering the details [13–18]. Most efforts for saliency detection are devoted to the design of advanced network architectures for multiscale learning. Although some methods can determine the salient object's position, they still have defects in generating maps with clear boundaries, as shown in Fig. 1.

Some existing methods in SOD aim to integrate shallower features layer-by-layer [21–28]. By connecting the features at the corresponding level in the encoder to the decoder, only the scale specific information can by characterized by the single level features. Some methods combine the low-level features and high-level features directly, which may produce noise or induce performance degradation [21, 29–35]. Some approaches combine the features from multiple layers in a fully-connected way to utilize the multi-level features [13, 16, 22, 36–38]. However, combining plethora features may lead to high computational cost and plenty of noise, thereby degrade the performance. For further performance gain, some methods employ a specific branch or an additional network [24, 26]. Netherless, these methods are not conducive for further applications because of computational redundancy and training difficulties.

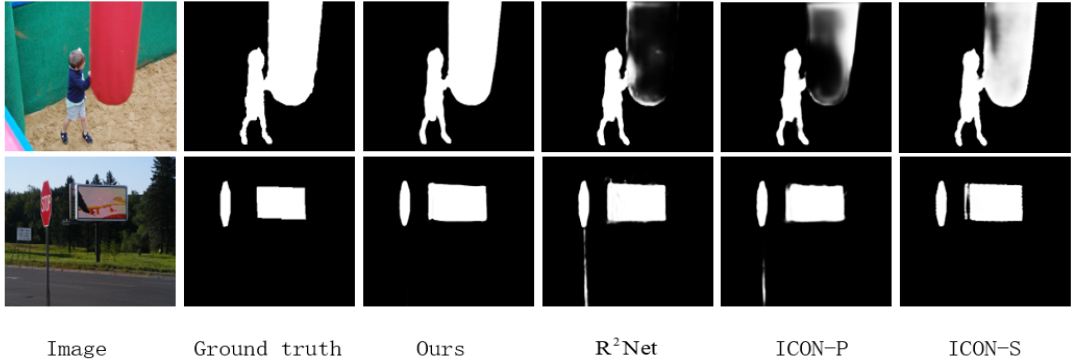

Fig. 1. Comparison of our method with $R^2$Net [19], ICONP [20], and ICON-S [20] for sample examples

    To break through the bottleneck of SOD, we propose a novel multistep feature aggregation network, namely MSFA, to investigate the other aspects of multi-scale learning. In particular, we use Transformer [39] as our backbone for feature information extraction. To enrich the semantic information of features, we design the Diverse Reception (DR) module, which stack maxpool operations with different kernels. To make better use of multi-level features and effectively integrate the contextual information from adjacent resolutions, we design a Multiscale Interaction (MSI) module. This module is composed of two parts, Composite Feature Integration (CFI) and Feature Decoder (FD) where mutual learning mechanism is incorporated into feature extraction. They consider the global structure and sample imbalance problem can be ameliorated. In such a way, low-level features for saliency details and accurate boundaries, and high-level features for contextual semantics can be fused more directly and efficiently. In order to refine the features, we introduce Feature Enhancement (FE) module to solve the holes problem caused by the fusion of varied layers in the saliency prediction.

    Our contributions can be summarized as follows:

・We propose a diverse reception module to effectively meet the scale challenges of features, which can obtain low-level detailed information through small kernel maxpool, high-level semantic information by middle kernel maxpool and global contextual information with large kernel maxpool.

・We design a multiscale interaction module to efficiently utilize the features from adjacent layers by mutual learning, which makes the network adaptively extract varied information from the data and better locating salient regions.

・We compare the proposed method with 3 state-of-the-art methods on six datasets. The results demonstrate the effectiveness and superiority of the proposed method under different evaluation metrics.

**2. Related works**

    In recent years, a plethora of methods in the field of salient object detection have been proposed and achieved encouraging performance for various benchmark datasets. This section will introduce some influential related works in the literature.

2.1. Salient Object Detection

    The initiated works were mainly based on hand-crafted features, and many low-level learning methods have been designed [7, 9–12]. And heuristic saliency priors were heavily used, such as color contrast, background prior and center prior, etc. Due to their limited feature representation capability, their generalization and effectiveness are limited especially compared with recently proposed methods. With the vast successes achieved by convolutional neural networks in other computer vision tasks, more and more researchers tried to use deep-learning based methods for SOD. Early CNN-based methods process and identify each image region for saliency prediction [40–42]. These methods have low computational efficiency and discard the spatial layout of the input image.

    By designing advanced network architectures for multi-level feature fusion, many recent CNN-based methods can

achieve promising results. Zhang et al. [29] simply combined all level features, thus easily produced information redundancy and noise interference because of coarse fusion. Hou et al. [43] integrated levels of features by constructing short connection structure, and obtained more accurate saliency maps. Wei et al. [44] proposed label decoupling framework, such that a saliency label is decomposed into a body map and detail map. The body map focuses on the pixels far away from edges, and the detail map concentrates on both edges and nearby pixels. Their network produced more precise saliency maps by making full use of complementary information between two branches. Zhuge et at. [20] proposed ICON method where they defined integrity learning mechanism at two levels. At the micro level, the model concentrates on part-whole relationship within a single salient object. At the macro level, the model identifies all salient objects within a given image. They proposed transformer-based architectures for context learning problem, which achieved prominent performance.

Effective loss can prompt the model to highlight the foreground region as smoothly as possible and deal well with the sample imbalance issue, therefore some works also tried to construct effective loss functions. Pang et at. [18] proposed the MINet to make better use of multi-level features and avoid the interference in feature fusion. Their method can efficiently utilize the features from adjacent layers through mutual learning, such that the network can adaptively extract multiscale information and better deal with scale variation. They proposed consistency enhanced loss (CEL) to address the fore-/back-ground imbalance issue caused by various scales. This loss can help solve the intra-class inconsistency and inter-class indistinction issues, thereby promoting the predicted boundaries to become sharper. Zhang et at. [19] proposed $R^2$Net for salient object detection, which can obtain more comprehensive high-level semantic information through dilated convolution and retain low-level detailed information with residual fusion. Their network can capture the relationship between different salient regions and improve the completeness of generated saliency maps. They designed a structural polishing loss (SP) to enable the network draw attention to the boundary details and produced saliency maps with finer boundaries.

**3. Methodology**

3.1. Overview

To deal with the prevalent scale variation issue in salient object detection, we propose a multistep feature aggregation (MSFA) network in this paper, and an overview of our architecture is illustrated in Fig. 2. We use the SwinTransformer [39] as the backbone to encode the image to five layers of features, and the last four layers are used. Then the following are the Diverse Reception (DR), Multiscale interaction (MSI) and Feature Enhancement (FE) for contextual feature fusion and saliency prediction.

3.2. Encoder

The encoder uses the pre-trained SwinTransformer [39] as the backbone to extract multi-level features which has five convolution stages. Usually, pre-trained RestNet [45] or VGG-16 [46] are used as the backbone network in SOD, here we use SwinTransformer. For simplicity, we denote the features generated by the backbone as $\{bf_1, bf_2, bf_3, bf_4, bf_5\}$. Assume that the size of the input image is $384 \times 384 \times 3$, then $bf_i$ has size $[\frac{384}{2^{i+1}}, \frac{384}{2^{i+1}}]$ for $i = 1,2,3,4$, and $[\frac{384}{2^i}, \frac{384}{2^i}]$ for $i = 5$. The shallower features with larger resolutions contain much spatial structural details, and the deeper features with smaller resolutions contain more coarse localization of salient objects. Like many other recent works [16, 17, 44, 47], we don't use the first layer feature in the decoder for its large spatial size causing much computational cost.

3.3. Diverse reception module

Enriching the receptive fields of the convolution can help the network learn features with scale and shape variety. This work uses maxpool operation with different kernel sizes, set to 3,7 and 11 for the feature $bf_i$ to enhance the diversity of extracted multi-level information. To preserve the rich details of the original features, we concatenate them with $bf_i$. As shown in Fig. 3, the whole process of the DR module can be described as the following formulas:

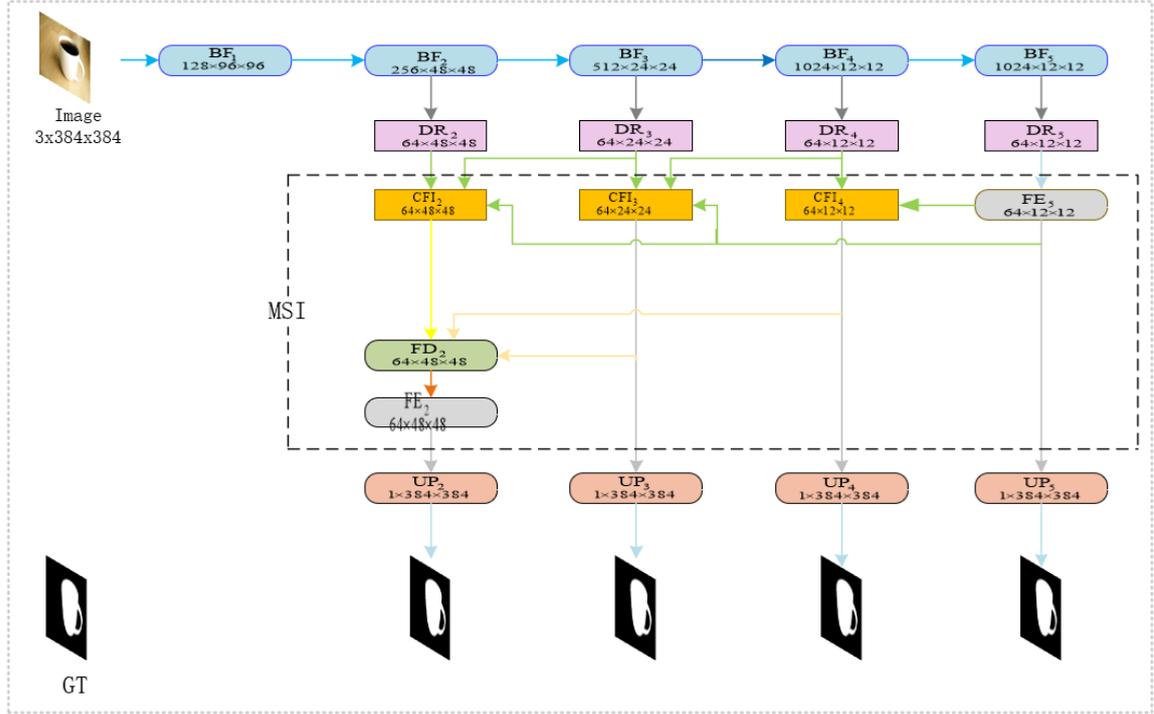

**Fig. 2.** The overall architecture of the proposed MSFA. Our model takes a RGB image (3 × 384 × 384) as input, and exploits the SwinTransformer [39] to extract multi-level features. Global and detail features are obtained by DR. They are combined by MSI (which consists of CFI and FD), then FE further refines the features. The final predictions are supervised by the ground truth.

$$bf_{ij} = Max_j(bf_i) \quad \text{for} \quad j = 3, 7, 11 \tag{1}$$

$$dr_ic = Concat(bf_{i3}, bf_{i7}, bf_{i11}, bf_i) \tag{2}$$

$$dr_i = Conv(dr_ic) \tag{3}$$

where $Max_j(\cdot)$ represents MaxPool operation with kernel size j, $Concat(\cdot)$ denotes channel-wise concatenation and $Conv(\cdot)$ means $3 \times 3$ convolution followed by batch normalization and Relu layers.

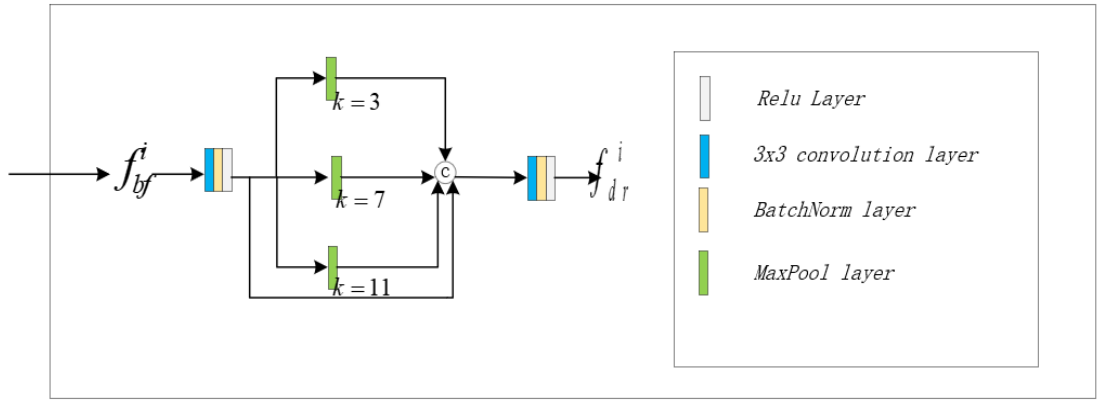

**Fig. 3.** Diverse reception module

3.4. Multiscale interaction module

This process can be considered as the combination of two successive submodules: Composite Feature Integration (CFI)

and Feature Decoder (FD) as shown in Fig. 4. The input of MSI consists of four features $f_{dr}^2, f_{dr}^3, f_{dr}^4$ and $f_{dr}^5$ which are the outputs of the diverse reception module with $f_{dr}^5$ enhanced. The entire process can be described in a two-step manner: first through CFI, then FD, which we will describe separately.

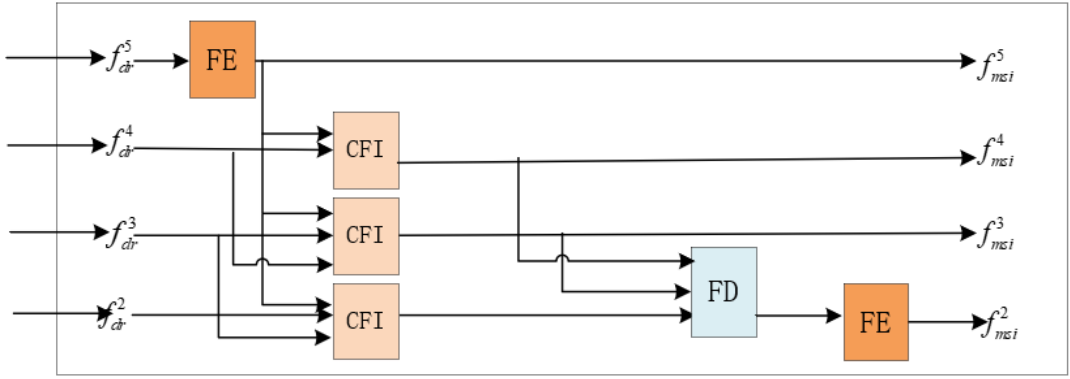

**Fig. 4**. Multiscale interaction module

For the composite feature integration, features $f_{dr}^i$, $f_{dr}^{i+1}$ and $f_{dr}^5$ are sent to the $i$-th CFI, with $f_{dr}^5$ enhanced. These features are aggregated by a strategy of interactive learning (Fig. 5). In many previous works, the context fusion of high-level features and low-level features after up-sampling are often finished with channel-wise concatenation or element-wise addition. Here, we adopt an efficient and straightforward operation, i.e., pixel-wise multiplication, which can suppress the background noise meanwhile strengthen the response of salient objects. The entire process can be formulated as follow:

$$f_{drc}^i = Conv_3(f_{dr}^i) \qquad (4)$$

$$f_h^i = f_{drc}^i \odot Conv_3(\text{UP}(f_{dr}^{i+1})) \qquad (5)$$

$$f_m^i = f_{drc}^i \odot Conv_3(\text{UP}(\text{FE}(f_{dr}^5))) \qquad (6)$$

$$f_l^i = Conv_3(\text{UP}(f_{dr}^{i+1})) \odot Conv_3(\text{UP}(\text{FE}(f_{dr}^5))) \qquad (7)$$

$$f_{cfi}^i = \text{FE}(Conv_3(Concat(f_{drc}^i, f_h^i, f_m^i, f_l^i))) \qquad (8)$$

where $Conv_3(\cdot)$ is a $3 \times 3$ convolution layer with a Batchnorm layer and Relu layer, and FE is the feature enhancement module that will be explained in the next section.

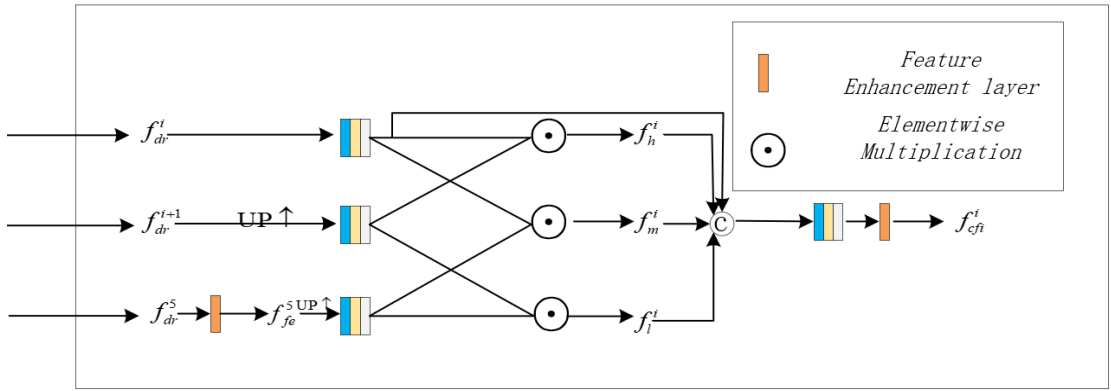

**Fig. 5.** Composite feature integration module

One-step feature aggregation methods may perform not well in some complex scenes and fail to completely detect the salient regions due to the complicated relationship among different parts of salient object or multiple salient objects, especially in challenging scenarios such as cluttered background, foreground disturbance and multiple salient objects. Thus, succeeding the above submodule CFI, we propose the submodule Feature Decoder (FD). In this case, features $f_{cfi}^2$, $f_{cfi}^3$ and $f_{cfi}^4$ strengthened by FE are fused using pixel-wise addition as usually done in previous works (Fig. 6). The process can be described as

$$f_{fdc}^3 = Conv_3(\text{UP}(f_{cfi}^4)) \oplus f_{cfi}^3 \tag{9}$$

$$f_{fd}^2 = Conv_3(Conv_3(\text{UP}(f_{fdc}^3)) \oplus f_{cfi}^2) \tag{10}$$

where $Conv_3(\cdot)$ is a $3 \times 3$ convolution layer with a Batchnorm layer and Relu layer.

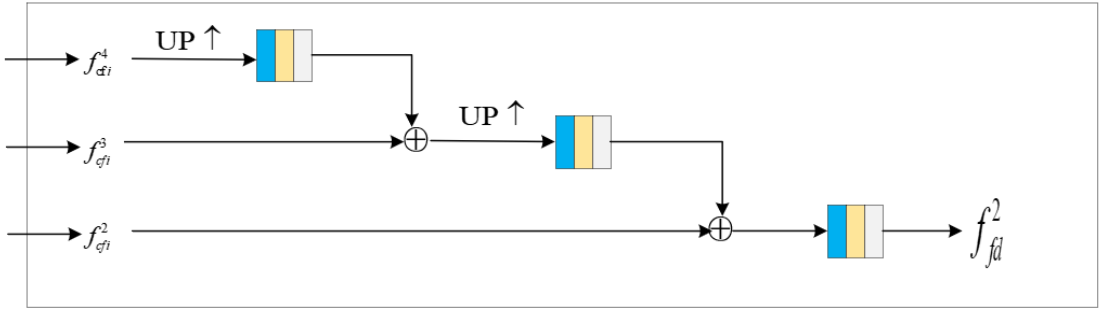

**Fig. 6.** Feature decoder module

3.5. Feature enhancement module

Through MSI module, comprehensive feature information can be obtained by combining the complementary characteristics between multilayer features. But there are some defects caused by the contradictory interweaving of different layers. Hence, we design FE to optimize and enhance the feature maps. For the input feature $f_{fd}^i$, we first use $Conv_3(\cdot)$ to obtain finer representation. Then, a $3 \times 3$ convolution layer is applied to expand the channel to 128 which will be divided into two parts each with channel size 64, generating $w_{fd}^i$ for multiplication and $b_{fd}^i$ for addition. The main process can be seen in Fig. 7 and described as follows

$$f_{fdc}^i = Conv_3(f_{fd}^i) \tag{11}$$

$$w_{fd}^i, b_{fd}^i \leftarrow conv_1(f_{fdc}^i) \tag{12}$$

$$f_{fe}^i = \theta(w_{fd}^i \cdot f_{fdc}^i + b_{fd}^i) \tag{13}$$

where $Conv_1(\cdot)$ is a $3 \times 3$ convolution layer and $\theta$ is the Relu layer.

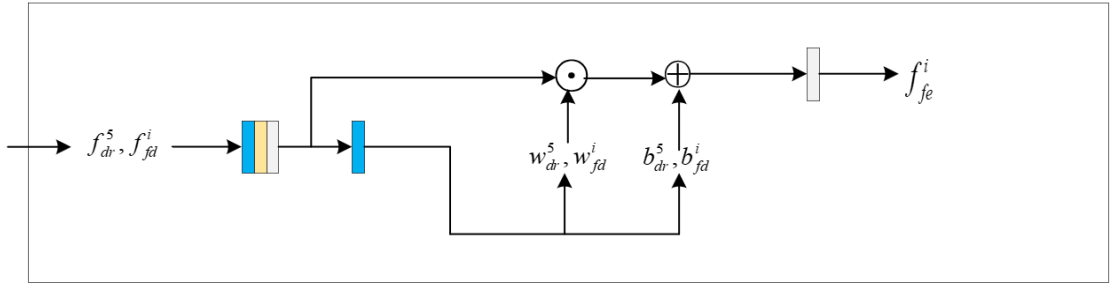

**Fig. 7.** Feature enhancement module

3.6. Loss function

In saliency detection, the widely used metric to measure the error between the ground truth and the prediction is the Binary Cross Entropy (BCE) loss, which can be formulated as

$$L_{bce} = -\frac{1}{H \times W}\sum_{i=1}^{H}\sum_{j=1}^{W}|G_{ij}\log P_{ij} + (1-G_{ij})\log(1-P_{ij})| \tag{14}$$

where $H$ and $W$ represent the height and width of the image, $G$ and $P$ denote the ground truth label and saliency prediction of the pixel $(i,j)$. The BCE may lead to poor performance in complex scenes, since it calculates pixel-wise difference separately, and cannot get a whole understanding of the image. To overcome this problem, the Intersection over Union (IoU) loss is introduced into the salient object detection task [18, 48], which can be formulated as

$$L_{iou} = 1 - \frac{\sum_{i=1}^{H}\sum_{j=1}^{W} G_{ij} \times P_{ij}}{\sum_{i=1}^{H}\sum_{j=1}^{W} G_{ij} + P_{ij} - G_{ij} \times P_{ij}} \tag{15}$$

In order to improve the boundary of the predicted map, Boundary (Bd) loss proposed in [49] will be used. Let

$$P^b = 1 - \frac{\sum_{i=1}^{H}\sum_{j=1}^{W} G_{ij}^b \times P_{ij}^b}{\sum_{i=1}^{H}\sum_{j=1}^{W} P_{ij}^b} \tag{16}$$

and

$$R^b = 1 - \frac{\sum_{i=1}^{H}\sum_{j=1}^{W} G_{ij}^b \times P_{ij}^b}{\sum_{i=1}^{H}\sum_{j=1}^{W} G_{ij}^b} \tag{17}$$

be the precision and recall calculated from the boundary of the ground truth and saliency map, denoted by $G^b$ and $P^b$ respectively. The boundary loss can be formulated as

$$L_{bd} = 1 - \frac{2 \times P^b \times R^b}{P^b + R^b} \tag{18}$$

Thus the loss function in our method can be formulated as

$$L = L_{bce} + L_{iou} + L_{bd} \tag{19}$$

The outputs of our MSFA are the saliency maps $m_i (i = 2,3,4,5)$. The sum of the loss of these saliency maps with the ground truth is the total loss. In addition, $m_2$ is the model result, providing the dominant loss, which is smaller than the auxiliary loss corresponding to the other saliency maps. Different weights are assigned, and the total loss is

$$L_{total} = L_{dom} + \sum_{i=3}^{5} \frac{1}{2^{i-2}} L_{aux}^i \tag{20}$$

**4. Experiments**

In this section, benchmark datasets, evaluation metrics and implementation details are described. Experiments are conducted on these datasets to compare our method with three state-of-the-art methods to analyze the effectiveness of the proposed model. For ResNet based models, since $R^2Net$ [19] is the best-known method, which performs favorably better than others (including VGG based methods), only $R^2Net$ is included for comparison. The other two models, ICON-P and ICON-S [20] are transformer based.

4.1. Datasets

We train our MSFA on six benchmark datasets commonly used for the SOD task, including DUTS, ECSSD, HKU-IS, DUT-OMRON, PASCAL-S, and SOC. DUTS is currently the largest saliency dataset, consisting of 15,572 images. Among them, DUTS-TR contains 10,553 images for training and DUTS-TE contains 5019 images for testing. ECSSD consists of 1000 images manually collected from the internet, that are semantically meaningful but with complex structures. HKU-IS includes 4447 images, many of which have multiple salient objects with low color contrast. DUT-OMRON has 5168 images with relatively cluttered background and one or more salient objects, thus this dataset is very challenging. PASCAL-S contains 850 images carefully selected from the PASCAL VOC dataset. SOC contains images with complicated backgrounds, which are more challenging than the other datasets. This dataset has 3600 images for training, 1200 images for validation and 1200 images for test. They are divided into 9 sub-datasets, corresponding to 9 different types of complex scenes.

4.2. Evaluation metrics

To quantitatively evaluate the performance of our model, we adopt 7 different evaluation metrics, including Mean Absolute Error (MAE), F-measure ($F_\beta$), weighted F-measure ($F_\beta^\omega$), S-measure ($S_\beta$), E-measure ($E_\beta$), precision recall (PR) curves, and F-measure curves.

Let $H, W$ denote the height and width of the image, then the MAE is defined as

$$MAE = \frac{1}{H \times W} \sum_{i=1}^{H} \sum_{j=1}^{W} |G_{ij} - P_{ij}| \tag{21}$$

where $G$ and $P$ represent the ground truth and saliency map respectively. Thus, MAE measures the average pixel-wise difference between them. The smaller the value is, the better the model performs.

Set $\beta^2 = 0.3$ as suggested by previous works [21, 37, 50], F-measure is defined by

$$F_\beta = \frac{(1+\beta^2) \times \text{Precision} \times \text{Recall}}{\beta^2 \times \text{Precision} + \text{Recall}} \tag{22}$$

It is the weighted harmonic mean of recall and precision.

Let $\text{Precision}^\omega$ and $\text{Recall}^\omega$ be the weighted precision and recall, the weighted F-measure is as follows

$$F_\beta^\omega = \frac{(1+\beta^2) \times \text{Precision}^\omega \times \text{Recall}^\omega}{\beta^2 \times \text{Precision}^\omega + \text{Recall}^\omega} \tag{23}$$

As a widely adopted metric, $F_\beta^\omega$ is used to overcome inaccurate evaluation caused by "interpolation defects, dependency defects, and defects of equal importance" [51].

To evaluate the similarity between the prediction and the ground truth, E-measure combines the local pixel values with the image-level mean value in one term. This metric can be defined as

$$E_\xi = \frac{1}{H \times W} \sum_{i=1}^{H} \sum_{j=1}^{W} \theta(\xi) \tag{24}$$

where $\xi$ and $\theta(\xi)$ denote the alignment matrix and enhanced alignment matrix.

S-measure is used to evaluate the structural similarity of the predicted saliency map and ground truth, and is closer to human visual perception. Let $S_o$ and $S_r$ be the object-aware and region-aware perception structures, then this metric is defined as

$$S_m = (1-\alpha)S_r + \alpha S_o \tag{25}$$

where $\alpha$ is set to be 0.5 as suggested in previous works.

Precision and recall are calculated by comparing the binary saliency map with ground truth mask pixel by pixel. For this, the binarized saliency map is generated by varying the threshold from 0 to 255, then (Precision, Recall) pairs and (F-measure, threshold) pairs are calculated to plot the PR curve and the F-measure curve. These metrics are essential to evaluate the performance of the model in SOD.

**Table I**

Performance comparison of the proposed method with state-of-the-art methods on UDTS, ECSSD and HKU-IS datasets. ↑ indicates that the higher the evaluation metric, the better the model effects and ↓ indicates that the lower the evaluation metric, the better the model effects. The best result is highlighted in red, and the second-best result is highlighted in blue.

| Dataset | DUTS | | | | | ECSSD | | | | | HKU-IS | | | | |
|---|---|---|---|---|---|---|---|---|---|---|---|---|---|---|---|
| Method | MAE↓ | $F_\beta$↑ | $\omega F_\beta$↑ | $E_m$↑ | $S_m$↑ | MAE↓ | $F_\beta$↑ | $\omega F_\beta$↑ | $E_m$↑ | $S_m$↑ | MAE↓ | $F_\beta$↑ | $\omega F_\beta$↑ | $E_m$↑ | $S_m$↑ |
| R$^2$Net | 0.031 | 0.869 | 0.864 | 0.916 | 0.899 | 0.029 | 0.936 | 0.925 | 0.931 | 0.929 | 0.025 | 0.924 | 0.915 | 0.958 | 0.924 |
| ICON-P | 0.026 | 0.868 | 0.882 | 0.919 | 0.917 | 0.024 | 0.936 | 0.933 | 0.928 | 0.940 | 0.022 | 0.925 | 0.925 | 0.963 | 0.935 |
| ICON-S | 0.025 | 0.882 | 0.886 | 0.930 | 0.917 | 0.023 | 0.941 | 0.936 | 0.932 | 0.941 | 0.022 | 0.928 | 0.925 | 0.965 | 0.935 |
| Ours | 0.023 | 0.889 | 0.889 | 0.929 | 0.914 | 0.021 | 0.948 | 0.943 | 0.936 | 0.943 | 0.020 | 0.934 | 0.932 | 0.967 | 0.936 |

4.3. Implementation details

Following most existing SOTA salient object detection methods, DUTS-TR is used as the training dataset. To avoid overfitting problem, random flip, random crop and multi-scale strategy are used as data augment techniques in the training stage. All experiments are runed on the publicly available Pytorch platform. To ensure model convergence, our network is trained on NVIDIA GTX 3090 GPU for 60 epochs with a batch size of 64. The backbone network is initialized with the pretrained SwinTranformer [20], and the rest parameters are initialized by the default setting of PyTorch. The momentum SGD optimizer with weight decay 5e-4, momentum 0.9 is used to update the model. To make the model more stable and accelerate the convergence speed in the training process, the learning rate warm-up strategy is used. Thus, the learning rate gradually increases from 1.6e-4 to 5e-3, and then gradually decreases to 1.6e-4. During testing, the images are resized to 384_384 to predict saliency maps without any post-processing.

**Table 2**

Performance comparison of the proposed method with state-of-the-art methods on DUT-OMRON and PASCAL-S datasets. ↑ indicates that the higher the evaluation metric, the better the model effects and ↓ indicates that the lower the evaluation metric, the better the model effects. The best result is highlighted in red, and the second-best result is highlighted in blue.

| Dataset | DUT-OMRON | | | | | PASCAL-S | | | | |
|---|---|---|---|---|---|---|---|---|---|---|
| Method | MAE↓ | $F_\beta$ ↑ | $\omega F_\beta$ ↑ | $E_m$ ↑ | $S_m$ ↑ | MAE↓ | $F_\beta$ ↑ | $\omega F_\beta$ ↑ | $E_m$ ↑ | $S_m$ ↑ |
| R$^2$Net | 0.045 | 0.793 | 0.777 | 0.878 | 0.844 | 0.057 | 0.853 | 0.834 | 0.864 | 0.860 |
| ICON-P | 0.047 | 0.794 | 0.793 | 0.886 | 0.865 | 0.051 | 0.854 | 0.847 | 0.865 | 0.875 |
| ICON-S | 0.042 | 0.811 | 0.804 | 0.897 | 0.869 | 0.048 | 0.865 | 0.854 | 0.869 | 0.878 |
| Ours | 0.039 | 0.817 | 0.808 | 0.898 | 0.867 | 0.046 | 0.871 | 0.859 | 0.872 | 0.874 |

**Table 3**

Performance comparison of the proposed method with state-of-the-art methods on SOC dataset. ↑ indicates that the higher the evaluation metric, the better the model effects and ↓ indicates that the lower the evaluation metric, the better the model effects. The best result is highlighted in red, and the second-best result is highlighted in blue.

| Dataset | AC | | | | | BO | | | | | CL | | | | |
|---|---|---|---|---|---|---|---|---|---|---|---|---|---|---|---|
| Method | MAE↓ | $F_\beta$ ↑ | $\omega F_\beta$ ↑ | $E_m$ ↑ | $S_m$ ↑ | MAE↓ | $F_\beta$ ↑ | $\omega F_\beta$ ↑ | $E_m$ ↑ | $S_m$ ↑ | MAE↓ | $F_\beta$ ↑ | $\omega F_\beta$ ↑ | $E_m$ ↑ | $S_m$ ↑ |
| R$^2$Net | 0.077 | 0.782 | 0.749 | 0.866 | 0.807 | 0.239 | 0.810 | 0.750 | 0.706 | 0.679 | 0.113 | 0.754 | 0.718 | 0.817 | 0.780 |
| ICON-P | 0.087 | 0.735 | 0.703 | 0.851 | 0.800 | 0.218 | 0.828 | 0.772 | 0.725 | 0.701 | 0.123 | 0.745 | 0.708 | 0.813 | 0.779 |
| ICON-S | 0.076 | 0.761 | 0.730 | 0.870 | 0.810 | 0.190 | 0.866 | 0.808 | 0.756 | 0.720 | 0.096 | 0.794 | 0.763 | 0.851 | 0.811 |
| Ours | 0.066 | 0.811 | 0.783 | 0.882 | 0.831 | 0.173 | 0.884 | 0.830 | 0.773 | 0.735 | 0.101 | 0.787 | 0.758 | 0.836 | 0.806 |
| Dataset | HO | | | | | MB | | | | | OC | | | | |
| Method | MAE↓ | $F_\beta$ ↑ | $\omega F_\beta$ ↑ | $E_m$ ↑ | $S_m$ ↑ | MAE↓ | $F_\beta$ ↑ | $\omega F_\beta$ ↑ | $E_m$ ↑ | $S_m$ ↑ | MAE↓ | $F_\beta$ ↑ | $\omega F_\beta$ ↑ | $E_m$ ↑ | $S_m$ ↑ |
| R$^2$Net | 0.088 | 0.805 | 0.770 | 0.863 | 0.819 | 0.066 | 0.802 | 0.772 | 0.887 | 0.832 | 0.096 | 0.751 | 0.710 | 0.839 | 0.781 |
| ICON-P | 0.108 | 0.738 | 0.702 | 0.827 | 0.793 | 0.107 | 0.681 | 0.641 | 0.805 | 0.766 | 0.124 | 0.695 | 0.653 | 0.794 | 0.755 |
| ICON-S | 0.092 | 0.782 | 0.748 | 0.855 | 0.817 | 0.095 | 0.743 | 0.698 | 0.841 | 0.792 | 0.103 | 0.749 | 0.706 | 0.829 | 0.783 |
| Ours | 0.081 | 0.820 | 0.794 | 0.868 | 0.835 | 0.078 | 0.790 | 0.763 | 0.873 | 0.819 | 0.087 | 0.771 | 0.737 | 0.844 | 0.799 |
| Dataset | HO | | | | | MB | | | | | OC | | | | |
| Method | MAE↓ | $F_\beta$ ↑ | $\omega F_\beta$ ↑ | $E_m$ ↑ | $S_m$ ↑ | MAE↓ | $F_\beta$ ↑ | $\omega F_\beta$ ↑ | $E_m$ ↑ | $S_m$ ↑ | MAE↓ | $F_\beta$ ↑ | $\omega F_\beta$ ↑ | $E_m$ ↑ | $S_m$ ↑ |
| R$^2$Net | 0.113 | 0.798 | 0.754 | 0.836 | 0.782 | 0.076 | 0.759 | 0.730 | 0.870 | 0.814 | 0.071 | 0.713 | 0.677 | 0.844 | 0.786 |
| ICON-P | 0.129 | 0.768 | 0.725 | 0.819 | 0.768 | 0.093 | 0.660 | 0.629 | 0.816 | 0.767 | 0.100 | 0.619 | 0.578 | 0.783 | 0.737 |
| ICON-S | 0.120 | 0.791 | 0.748 | 0.827 | 0.779 | 0.082 | 0.702 | 0.668 | 0.850 | 0.787 | 0.083 | 0.675 | 0.638 | 0.827 | 0.768 |
| Ours | 0.108 | 0.805 | 0.768 | 0.840 | 0.791 | 0.062 | 0.774 | 0.752 | 0.880 | 0.830 | 0.071 | 0.708 | 0.682 | 0.835 | 0.787 |

4.4. Comparison with state-of-the-art methods

We compare our method with three state-of-the-art saliency detection methods, namely ICON-P, ICON-S and R$^2$Net [19, 20]. Among them, ICON-P and ICON-S provide saliency maps with transformer as the backbone network, and R$^2$Net with ResNet-50 as the backbone network. For a fair comparison, all saliency maps of these methods are provided by the authors or generated by their released codes.

4.4.1 Quantitative comparison

For the metrics MAE, F-measure, weighted F-measure, E-measure, and S-measure, Table 1 and Table 2 provide the quantitative comparisons between our method and the other three methods on DUTS-TE, ECSSD, HKU-IS, DUT-OMRON, and PASCAL-S datasets. It shows that our MSFA achieves the best results in almost all metrics. Especially, the results of our method in MAE are reduced by 8.7%; 7.1%; 9.5%; 10%, and 4.3% respectively, compared with the suboptimal results on the DUT, DUT-OMRON, ECSSD, HKU-IS, and PASCAL-S datasets. For cases our method cannot obtain the best results, they are only about less than 0.5% compared with the best results. In general, the comparison results demonstrate the efficacy of our MSFA.

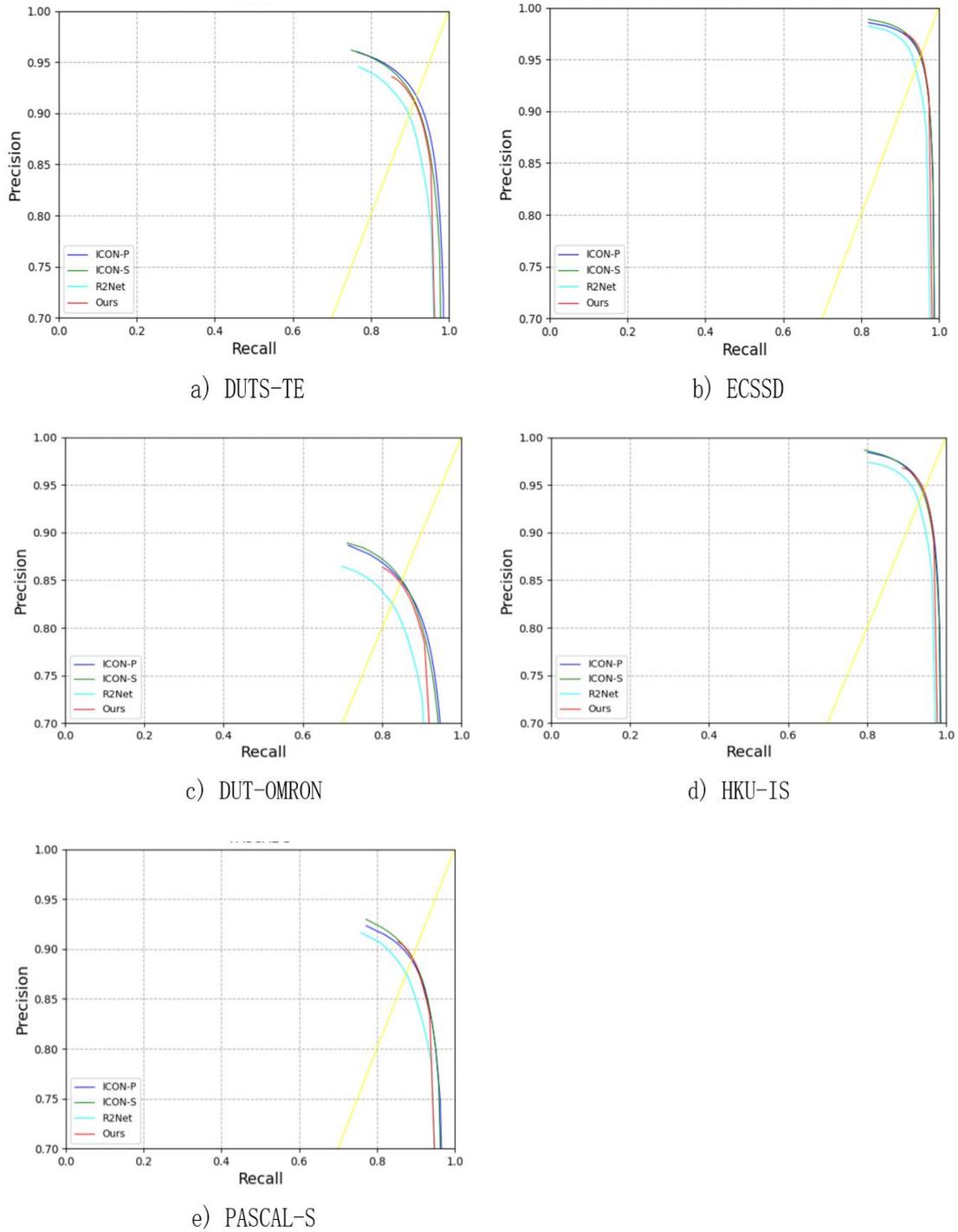

a) DUTS-TE

b) ECSSD

c) DUT-OMRON

d) HKU-IS

e) PASCAL-S

**Fig. 8.** PR curves of the proposed method and other three state-of-the-art methods

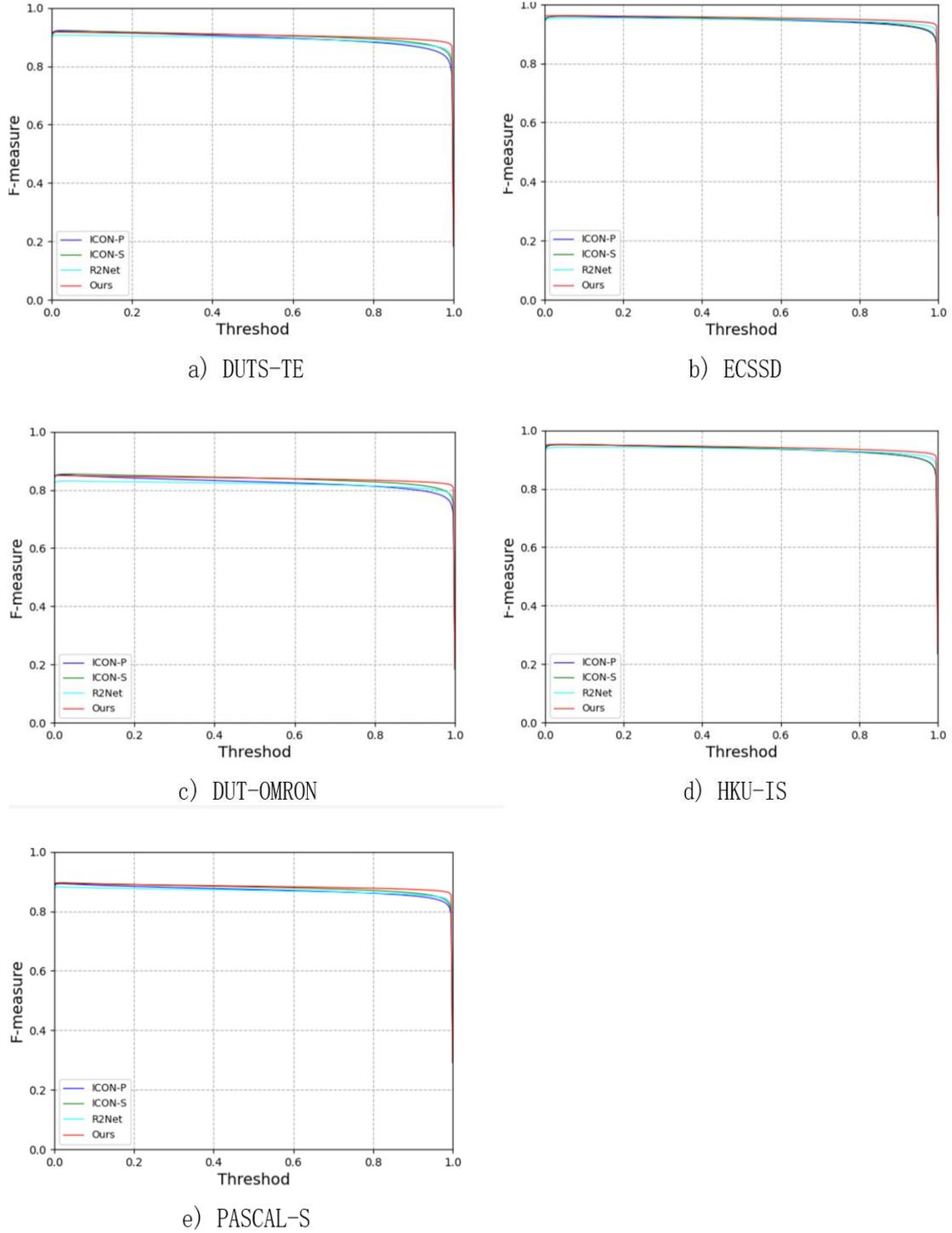

**Fig. 9.** F-measure curves of the proposed method and other three state-of-the-art methods

The SOC dataset contains nine different sub-datasets, AC(appearance change), BO(big object), CL(clutter), HO(heterogeneous object), MB(motion blur), OC(occlusion), OV(out-of-view), SC(shape complexity), and SO(small object). In Table 3, we compare our method with the other three methods for MAE, F-measure, weighted F-measure, E-measure, and S-measure on the nine sub-datasets. As seen, our method performs better than the other methods on most sub-datasets except for CL (ICON-S the best) and MB ($R^2$Net the best) , because of the diversity and different detection capability in complex scenarios.

In addition to the above results, we also present the PR curves and F-measure curves of these methods on the DUTS-TE,

ECSSD, HKU-IS, DUT-MORON, and PASCAL-S datasets in Fig. 8 and Fig. 9. For the PR curves, it can be seen that our method is comparable or locate higher than the other methods on ECSSD, HKU-IS, and PASCAL-S datasets, which can also be checked by the breakeven point (intersection with the yellow line). For the datasets DUTS-TE and DUT-MORON, the break-even points of our method seem lie slightly below the methods ICON-P and ICON-S. But, if we project the PR curves to the horizontal line, it can be seen that, our method has a shorter segment which lies to the right and more closer to 1. This means that our model generally always have higher Recall values than others, which is defined by

$$\text{Recall} = \frac{TP}{TP+FN} \tag{26}$$

showing that our method performs excellently, here TP means True Positives and FN represents False Negatives. For the F-measure curves, as shown in Fig. 9, our method is comparable or locate consistently higher than the other methods, demonstrating the robustness of the proposed model.

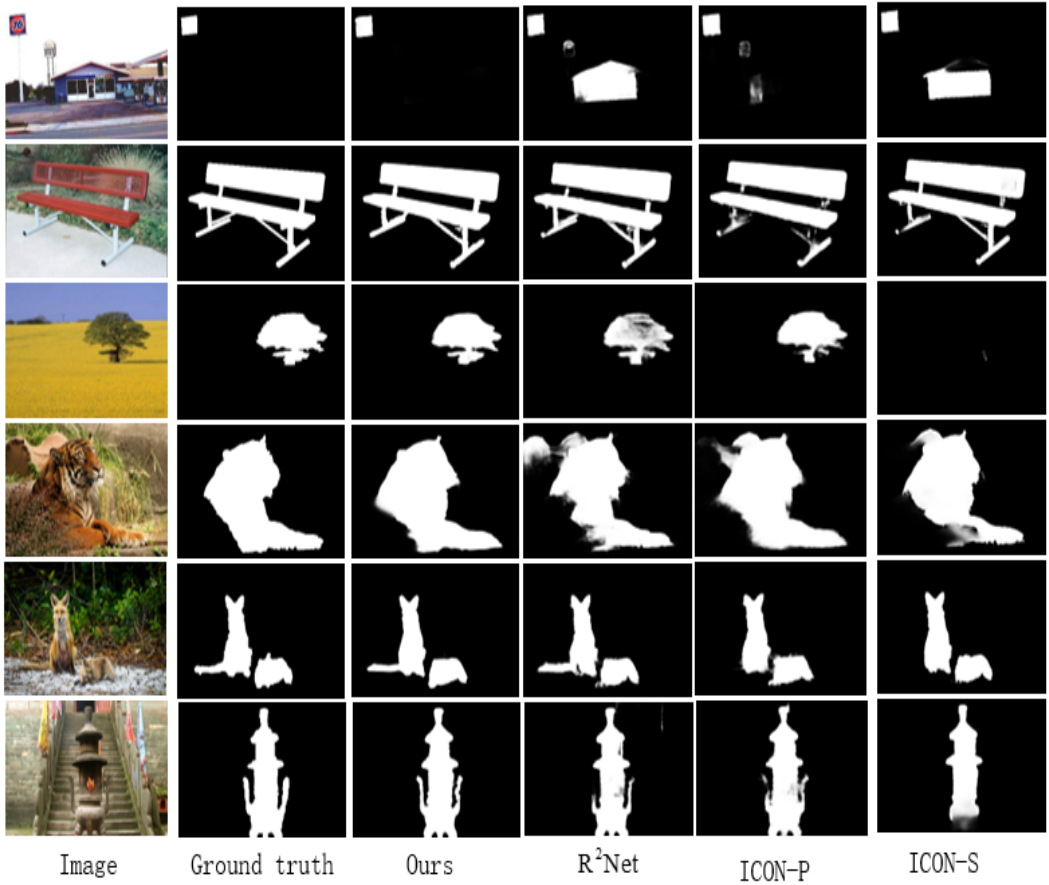

**Fig. 10.** Visual comparison results of our MSFA and other SOTA methods

4.4.2. Visual comparison

Some representative examples are provided in Fig. 10 for visual comparisons. These results include various scenarios, such as images with large size objects (lines 2,4), image with small objects (line 1), image with multiple salient objects (line 5), and image with low contrast between objects and background (line 4), and images with complex parts (lines 3,6). It can be seen that our proposed method can provide more accurate and complete saliency maps with clear boundaries.

4.5. Ablation study

In the section we conduct ablation study for the loss function and proposed modules through experiments. comparisons are based on the ECSSD and HKU-IS datasets in terms of MAE, F-measure, weighted F-measure, E-measure, and S-measure.

4.5.1. Effectiveness of the loss function

While keeping the model structure and training strategy unchanged, we use different losses to supervise the model, and the results are shown in Table 4. It can be seen that the model trained with BCE+IoU+Bd can achieve consistent performance enhancement in terms of all five metrics. As shown in Fig. 11, the saliency maps produced with the composite of the three losses as supervision have finer boundaries.

Table 4

Performance comparison of MSFA with different loss functions on ECSSD and HKU-IS datasets. ↑ indicates that the higher the evaluation metric, the better the model effects and ↓ indicates that the lower the evaluation metric, the better the model effects. The best result is highlighted in red.

| Dataset | ECSSD | | | | | HKU-IS | | | | |
|---|---|---|---|---|---|---|---|---|---|---|
| Loss function(s) | MAE↓ | $F_\beta$ ↑ | $\omega F_\beta$ ↑ | $E_m$ ↑ | $S_m$ ↑ | MAE↓ | $F_\beta$ ↑ | $\omega F_\beta$ ↑ | $E_m$ ↑ | $S_m$ ↑ |
| BCE | 0.028 | 0.927 | 0.919 | 0.926 | 0.940 | 0.026 | 0.909 | 0.904 | 0.958 | 0.934 |
| BCE+IoU | 0.024 | 0.941 | 0.936 | 0.932 | 0.943 | 0.021 | 0.928 | 0.926 | 0.966 | 0.937 |
| BCE+IoU+Bd | 0.021 | 0.948 | 0.943 | 0.936 | 0.943 | 0.020 | 0.934 | 0.932 | 0.967 | 0.936 |

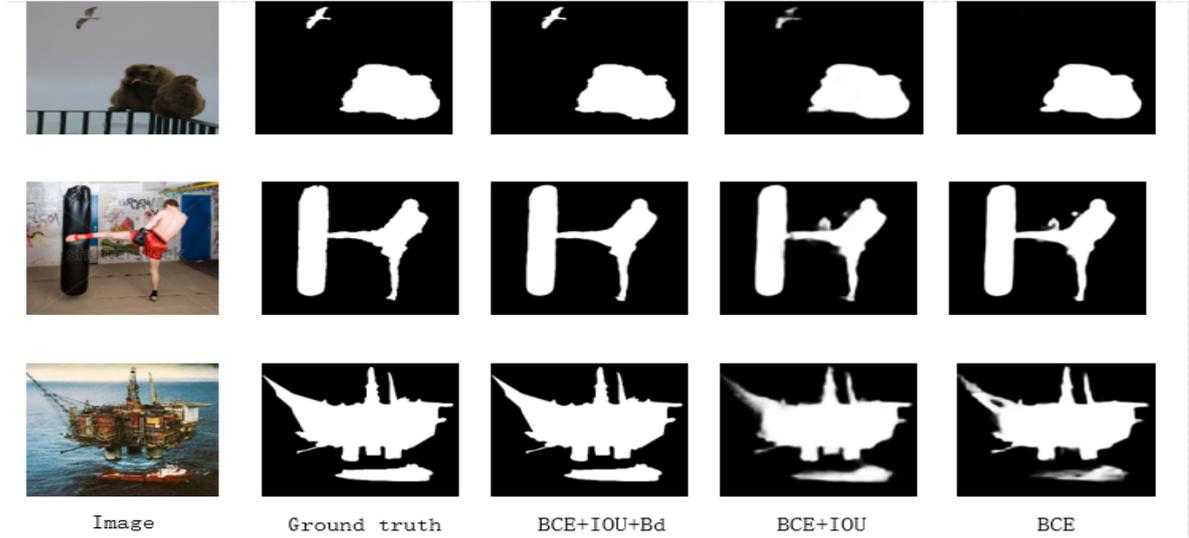

Fig. 11. Visual comparison results of our MSFA with different loss functions

Table 5

Performance comparison of proposed modules on ECSSD and HKU-IS datasets. ↑ indicates that the higher the evaluation metric, the better the model effects and ↓ indicates that the lower the evaluation metric, the better the model effects. The best result is highlighted in red.

| Dataset | ECSSD | | | | | HKU-IS | | | | |
|---|---|---|---|---|---|---|---|---|---|---|
| Method | MAE↓ | $F_\beta$ ↑ | $\omega F_\beta$ ↑ | $E_m$ ↑ | $S_m$ ↑ | MAE↓ | $F_\beta$ ↑ | $\omega F_\beta$ ↑ | $E_m$ ↑ | $S_m$ ↑ |
| BASE | 0.025 | 0.939 | 0.934 | 0.932 | 0.935 | 0.022 | 0.926 | 0.923 | 0.964 | 0.927 |
| BASEMSI | 0.023 | 0.944 | 0.939 | 0.932 | 0.940 | 0.020 | 0.933 | 0.930 | 0.966 | 0.934 |
| BASEMSI+DR | 0.022 | 0.946 | 0.942 | 0.932 | 0.942 | 0.020 | 0.934 | 0.931 | 0.967 | 0.935 |
| BASEMSI+DR+FE | 0.021 | 0.948 | 0.943 | 0.936 | 0.943 | 0.020 | 0.934 | 0.932 | 0.967 | 0.936 |

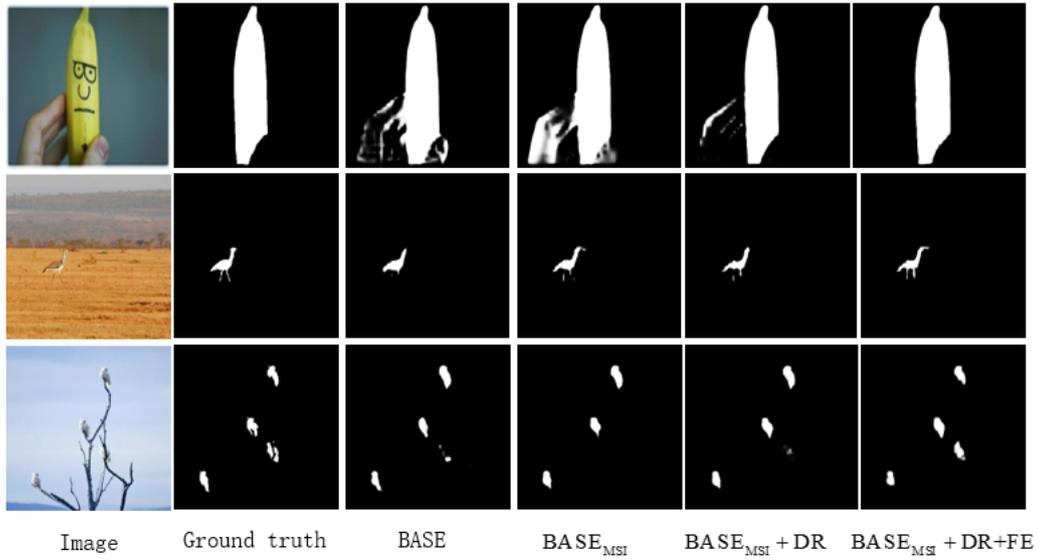

Fig. 12. Visual comparison results of proposed modules

4.5.2. Effectiveness of the proposed modules

Our proposed MSFA consists of three modules, namely the Diverse Reception (DR) module, Multiscale interaction (MSI) module, and Feature Enhancement (FE) module. In order to illustrate the effectiveness of these modules, we conduct a detailed analysis by keeping the training strategy unchanged, and adding these modules one by one. The base model contains SwinTransformer [20] as encoder and a basic decoder, where multi-level features from the encoder are concatenated channel-wise and compressed by convolution layer to a saliency map.    We change the base model by adding the proposed MSI to get  $BASE_{MSI}$. Then DR is added to get  $BASE_{MSI}$+DR, and finally FE is included to get $BASE_{MSI}$+DR+FE. As shown in Table 5, each component of our network help to improve the performance over the baseline model over all five metrics The visual effects of different modules are shown in Fig. 12. We can see that the proposed DR and MSI help effectively suppress the interference of backgrounds and obtain finer saliency detection results by capturing richer multi-scale contextual information. FE helps to improve the boundary details, making the saliency map more closer to the ground truth, suggesting the necessity of each proposed components.

5. Conclusion

In this paper we present a Multistep Feature Aggregation (MSFA) network to detect salient objects from given image scenes. First, we propose DR containing variable kernels to obtain comprehensive information for local and global features. After that, we design MSI to integrate multi-level features of adjacent layers and learn contextual knowledge from branches of different resolutions to boost the capability of detecting size-varying objects. Finally, FE is proposed to alleviate the impact of direct fusion of different levels of features. Combined supervision with three loss functions is used to produce high-quality saliency maps. Extensive experiments on six datasets demonstrate the state-of-the-art performance of the proposed model.

**Declaration of Competing Interest**

The manuscript has not been published before and is not being considered for publication elsewhere. All authors have contributed to the creation of this manuscript for important intellectual content, and read and approved the final manuscript. We declare there is no conflict of interest.

**Acknowledgements**

The authors would like to thank anonymous referees for any helpful suggestions and advice.